%% file: main.tex
\documentclass[runningheads]{llncs}

\usepackage{eccv}

\usepackage{eccvabbrv}

\usepackage{graphicx}
\usepackage{booktabs}

\usepackage[accsupp]{axessibility}  

\usepackage{hyperref}

\usepackage{orcidlink}
\usepackage{amssymb}
\usepackage{makecell}
\usepackage{pifont}
\newcommand{\cmark}{\ding{51}} 
\newcommand{\xmark}{\ding{55}} 
\usepackage{xcolor}
\newcommand{\gray}[1]{\textcolor{gray}{#1}}
\usepackage{multirow}
\usepackage{marvosym} 
\newcommand{\rowNumber}[1]{\textcolor{Cerulean}{#1}}

\begin{document}

\title{LaMI-DETR: Open-Vocabulary Detection\\ with Language Model Instruction} 


\author{
Penghui Du\inst{1,2,3}\thanks{Equal contribution. \textsuperscript{\Letter} Corresponding author} \and
Yu Wang\inst{2}$^\star$\and
Yifan Sun\inst{2} \and
Luting Wang\inst{1} \and
Yue Liao\inst{1} \and
Gang Zhang\inst{2} \and
Errui Ding\inst{2} \and
Yan Wang\inst{3}\textsuperscript{\Letter} \and
Jingdong Wang\inst{2} \and
Si Liu\inst{1}\textsuperscript{\Letter}
}

\authorrunning{
P. Du, Y. Wang et al.}

\institute{
${^1}$Beihang University ~~${^2}$Baidu ~~${^3}$AIR, Tsinghua University \\
\email{
$\{$dupenghui, wangluting, liusi$\}$@buaa.edu.cn ~~liaoyue.ai@gmail.com\\
~~$\{$wangyu106, sunyifan01, zhanggang03, dingerrui, wangjingdong$\}$@baidu.com ~~wangyan@air.tsinghua.edu.cn
}}

\maketitle

\input{sections/0_abstract}
\input{sections/1_intro}
\input{sections/2_relatedwork}

\input{sections/3_method/overview}
\input{sections/4_exp/overview}

\input{sections/5_conclusion}
\section*{Acknowledgements}
This research is supported in part by National Science and Technology Major Project (2022ZD0115502), National Natural Science Foundation of China (NO. 62122010, U23B2010), Zhejiang Provincial Natural Science Foundation of China (Grant No. LDT23F02022F02), and Beijing Natural Science Foundation (NO. L231011). We thank the authors of LW-DETR~\cite{chen2024lw}: Qiang Chen and Xinyu Zhang, the author of OADP~\cite{OADP}: Yi Liu and the author of DetPro~\cite{detpro}: Yu Du for their helpful discussions.

%
%
\bibliographystyle{splncs04}
\bibliography{main}
\clearpage
\input{sections/6_supp}
\end{document}

%% file: sections/0_abstract.tex
\begin{abstract}
    Existing methods enhance open-vocabulary object detection by leveraging the robust open-vocabulary recognition capabilities of Vision-Language Models (VLMs), such as CLIP. 
    However, two main challenges emerge: 
    (1) A deficiency in concept representation, where the category names in CLIP's text space lack textual and visual knowledge.
    (2) An overfitting tendency towards base categories, with the open vocabulary knowledge biased towards base categories during the transfer from VLMs to detectors.
    To address these challenges, we propose the Language Model Instruction (LaMI) strategy, which leverages the relationships between visual concepts and applies them within a simple yet effective DETR-like detector, termed LaMI-DETR. 
    LaMI utilizes GPT to construct visual concepts and employs T5 to investigate visual similarities across categories. 
    These inter-category relationships refine concept representation and avoid overfitting to base categories.
    Comprehensive experiments validate our approach's superior performance over existing methods in the same rigorous setting without reliance on external training resources.
    LaMI-DETR achieves a rare box AP of $43.4$ on OV-LVIS, surpassing the previous best by $7.8$ rare box AP.
    \keywords{Inter-category Relationships \and Language Model \and DETR}
\end{abstract}

%% file: sections/1_intro.tex
\section{Introduction}
\label{sec:intro}

Open-vocabulary object detection (OVOD) aims to identify and locate objects from a wide range of categories, including base and novel categories during inference, even though it is only trained on a limited set of base categories.
Existing works \cite{vild2021, detpro, zang2022open, wu2023cora, Shi_2023_ICCV, OADP, wu2023baron, kim2023contrastivecfm} in open-vocabulary object detection have been focusing on the development of sophisticated modules within detectors. 
These modules are tailored to effectively adapt the zero-shot and few-shot learning capabilities inherent in Vision-Language Models (VLMs) to the context of object detection.

However, there are two challenges in most existing methods:
(1) \textit{Concept Representation}. 
Most existing methods represent concepts using name embeddings from CLIP text encoder. However, this approach of concept representation has a limitation in capturing the textual and visual semantic similarities between categories, which could aid in discriminating visually confusable categories and exploring potential novel objects;
(2) \textit{Overfit to base categories}. Although VLMs can perform well on novel categories, only base detection data is used in open vocabulary detectors' optimization, resulting in detectors' overfitting to base categories. As a result, novel objects are easily regarded as background or base categories.

\begin{figure}[tb]
    \centering
    \includegraphics[width=\linewidth]{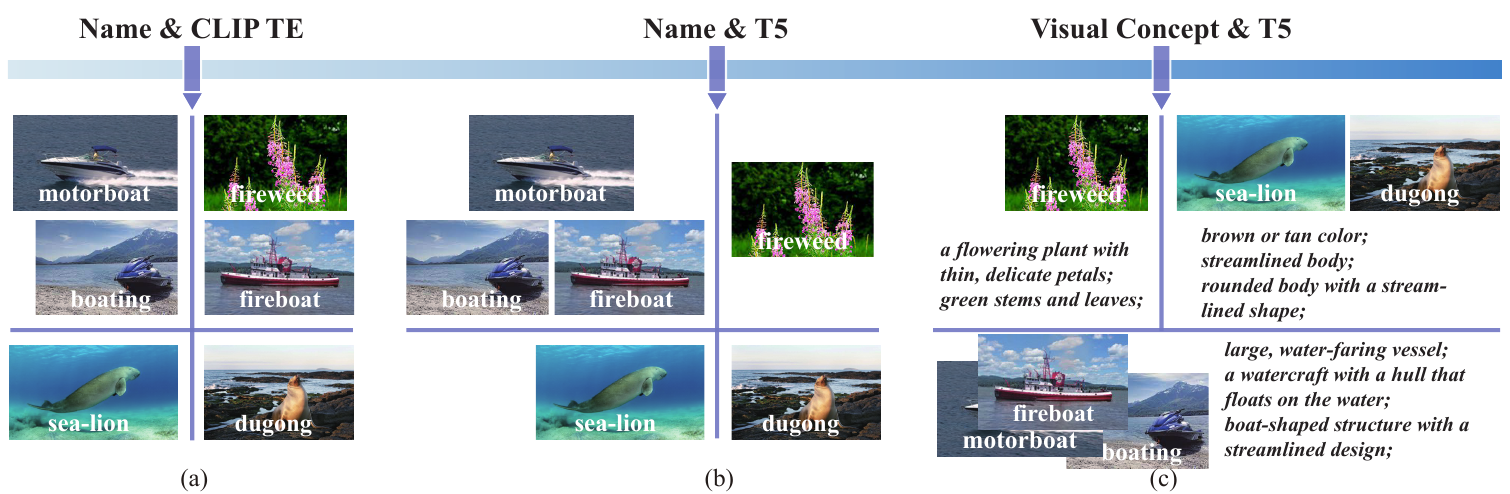}
    \caption{Illustration of the concept representation challenge. 
    The clustering results are from (a) name embeddings by CLIP text encoder, (b) name embeddings by T5, and (c) visual description embeddings by T5, respectively.
    (a) CLIP text encoder struggles to distinguish between category names that are compositionally similar in letters, such as "fireboat" and "fireweed".
    (b) T5 fails to cluster categories that are visually comparable but compositionally different in name around the same cluster center, such as "sea-lion" and "dugong".
    (c) Marrying T5's textual semantic knowledge with visual insights achieves reasonable cluster results.}

    \label{fig:cluster_c}
\end{figure}

Firstly, the issue of concept representation. 
Category names within CLIP's textual space are deficient in both textual depth and visual information.

(1) The VLM's text encoder lacks textual semantic knowledge compared with language model.
As depicted in Figure~\ref{fig:cluster_c}(a), relying solely on name representations from CLIP concentrates on the similarity of letter composition, neglecting the hierarchical and common-sense understanding behind language.
This method is disadvantageous for categorizing clustering as it fails to consider the conceptual relationships between categories.
(2) Existing concept representations based on abstract category names or definitions fail to account for visual characteristics.
Figure~\ref{fig:cluster_c}(b) demonstrates this problem, where sea lions and dugongs, despite their visual similarity, are allocated to separate clusters. 
Representing concept only with category name overlooks the rich visual context that language provides, which can facilitate the discovery of potential novel objects. 

Secondly, the issue of overfitting to base categories.
To leverage the open vocabulary capabilities of VLMs, we employ a frozen CLIP image encoder as the backbone and utilize category embeddings from the CLIP text encoder as classification weights. 
We regard that detector training should serve two main functions: firstly, to differentiate foreground from background; and secondly, to maintain the open vocabulary classification capability of CLIP. 
However, training solely on base category annotations, without incorporating additional strategies, often results in overfitting: novel objects are commonly misclassified as either background or base categories. This problem has been further elucidated in prior research~\cite{Shi_2023_ICCV, wang2023opencorpus}.

We pinpoint the exploration of inter-category relationships as pivotal in tackling the aforementioned challenges. 
By cultivating a nuanced understanding of these relationships, we can develop a concept representation method that integrates both textual and visual semantics. 
This approach can also identify visually similar categories, guiding the model to focus more on learning generalized foreground features and preventing overfitting to base categories.
Consequently, in this paper, we introduce \textbf{LaMI-DETR} (\textit{Frozen CLIP-based DETR with \textbf{La}nguage \textbf{M}odel \textbf{I}nstruction}), a simple but effective DETR-based detector that leverages language model insights to extract inter-category relationships, aiming to solve the aforementioned challenges.

To tackle the concept representation, we first adopt the Instructor Embedding~\cite{INSTRUCTOR}, a T5 language model, to re-evaluate category similarities. 
As we find that language models exhibit a more refined semantic space compared to the CLIP text encoder. 
As shown in Figure~\ref{fig:cluster_c}(b), "fireweed" and "fireboat" are categorized into separate clusters, mirroring human recognition more closely.
Next, we introduce the use of GPT-3.5~\cite{brown2020languagegpt3} to generate visual descriptions for each category. This includes detailing aspects such as shape, color, and size, effectively converting these categories into visual concepts. Figure~\ref{fig:cluster_c}(c) shows that, with similar visual descriptions, sea lions and dugongs are now grouped into the same cluster.
To mitigate the overfitting issue, we cluster visual concepts into groups based on visual description embeddings from T5. 
This clustering result enables the identification and sampling of negative classes that are visually different from ground truth categories in each iteration.
This relaxes the optimization of classification and focuses the model on deriving more generalized foreground features rather than overfitting to base categories.
Consequently, this approach enhances the model's generalizability by reducing overtraining on base categories while preserving CLIP image backbone's ability to categorize.

In summary, we introduce a novel approach, LaMI, to enhance base-to-novel generalization in OVOD.
LaMI harnesses large language models to extract inter-category relationships, utilizing this information to sample easy negative categories and avoid overfitting to base categories, while also refining concept representations to enable effective classification between visually similar categories.
We propose a simple but effective end-to-end LaMI-DETR framework, enabling the effective transfer of open vocabulary knowledge from pretrained VLMs to detectors.
We demonstrate the superiority of our LaMI-DETR framework through rigorous testing on large vocabulary OVOD benchmark, including $+7.8$ AP$_\textrm{r}$ on OV-LVIS and $+2.9$ AP$_\textrm{r}$ on VG-dedup(fair comparison with OWL~\cite{owlvit2023, owlvitv22023scaling}). 
Code is available at \url{https://github.com/eternaldolphin/LaMI-DETR}.

%% file: sections/2_relatedwork.tex
\section{Related Work}

\subsubsection{\textbf{Open-vocabulary object detection (OVOD)}}
leverages the image and language alignment knowledge stored in image-level dataset, \eg, Conceptual Captions~\cite{CC}, or large pre-trained VLMs, \eg, CLIP~\cite{clip}, to incorporate the open-vocabulary information into object detectors.
One group of OVOD utilizes large-scale image-text pairs to expand detection vocabulary~\cite{ovr-cnn, regionclip2022, detic2022,vlplm2022,ma2023codet,feng2022promptdet,Hanoona2022Bridging}
However, based on VLMs' proven strong zero-shot recognition abilities, most open-vocabulary object detectors leverage VLM-derived knowledge to handle open vocabularies. The methods for object detectors to obtain open vocabulary knowledge from VLM can be divided into three categories: pseudo labels~\cite{regionclip2022, Hanoona2022Bridging,zang2022open}, distillation~\cite{vild2021,detpro,OADP,wu2023baron} or parameter transfer~\cite{fvlm, wu2023cora}.
Despite its utility, performances of these methods are arguably restricted by the teacher VLM, which is shown to be largely unaware of inter-category visual relationship.
Our method is orthogonal to all the aforementioned approaches in the sense that it not only explicitly models region-word correspondences, but also leverages visual correspondences across categories to help localize novel categories, which greatly improves the performance, especially in the DETR-based architecture~\cite{zhang2022dino, hu2024dac, zhao2024ms, chen2024lw}.

\subsubsection{\textbf{Zero-shot object detection (ZSD)}} 
addresses the challenge of detecting novel, unseen classes by leveraging language features for generalization. Traditional approaches utilize word embeddings, such as GloVe~\cite{pennington2014glove}, as classifier weights to project region features into a pre-computed text embedding space~\cite{bansal2018zero, demirel2018zero}. This enables ZSD models to recognize unseen objects by their names during inference. 
However, the primary limitation of ZSD lies in its training on a constrained set of seen classes, failing to adequately align the vision and language feature spaces. 
Some methods attempt to mitigate this issue by generating feature representations of novel classes using Generative Adversarial Networks~\cite{goodfellow2014generative, zhao2020gtnet} 
or through data augmentation strategies for synthesizing unseen classes~\cite{zhu2020don}. 
Despite these efforts, ZSD still faces significant performance gaps compared to supervised detection methods, highlighting the difficulty in extending detection capabilities to entirely unseen objects without access to relevant resources.

\subsubsection{\textbf{Large Language Model (LLM)}}
Language data has increasingly played a pivotal role in open-vocabulary research, with recent Large Language Models (LLMs) showcasing vast knowledge applicable across various Natural Language Processing tasks. 
Works such as~\cite{menon2022visual, pratt2023doescupl, yang2023language} have leveraged language insights from LLMs to generate descriptive labels for visual categories, thus enriching VLMs without necessitating further training or labeling. 
Nonetheless, there are gaps in current methodologies: firstly, the potential of discriminative LLMs for enhancing VLMs is frequently overlooked; secondly, the inter-category relationships remain underexplored. 
We propose a novel, straightforward clustering approach that employs GPT and Instructor Embeddings to investigate visual similarities among concepts, addressing these oversights.

%% file: sections/3_method/overview.tex
\section{Method}

In this section, we begin with an introduction to open-vocabulary object detection (OVOD) in Section~\ref{sec:prelim}. 
Following this, we describe our proposed architecture of LaMI-DETR, a straightforward and efficient OVOD baseline, detailed in  Section~\ref{sec:fc_detr}. Finally, we provide a detailed explanation of Language Model Instruction (LaMI) in Section~\ref{sec:lami}. 

\subsection{Preliminaries}
\label{sec:prelim}
\input{sections/3_method/Preliminaries}

\subsection{Architecture of LaMI-DETR}
\label{sec:fc_detr}
\input{sections/3_method/overview_of_fcdetr}

\subsection{Language Model Instruction}
\label{sec:lami}
\input{sections/3_method/language_model_instruction}

%% file: sections/3_method/Preliminaries.tex
Given an image \(\mathbf{I} \in \mathbb{R}^{H \times W \times 3}\) as input to an open-vocabulary object detector, two primary outputs are typically generated:
(1) Classification, wherein a class label, \(c_j \in \mathcal{C}_{\text{test}}\), is assigned to the \(j^{\text{th}}\) predicted object in the image, with \(\mathcal{C}_{\text{test}}\) representing the set of categories targeted during inference.
(2) Localization, which involves determining the bounding box coordinates, \(\mathbf{b}_j \in \mathbb{R}^4\), that identify the location of the \(j^{\text{th}}\) predicted object.
Following the framework established by OVR-CNN~\cite{ovr-cnn}, there is a detection dataset, \(\mathcal{D}_{\text{det}}\), comprising bounding box coordinates, class labels, and corresponding images, and addressing a category vocabulary, \(\mathcal{C}_{\text{det}}\).

In line with the conventions of OVOD, we denote the category spaces of \(\mathcal{C}_{\text{test}}\) and \(\mathcal{C}_{\text{det}}\) as \(\mathcal{C}\) and \(\mathcal{C}_{\text{B}}\) respectively.
Typically, \(\mathcal{C}_{\text{B}} \subset \mathcal{C}\).
The categories within \(\mathcal{C}_{\text{B}}\) are known as base categories, whereas those exclusively appearing in \(\mathcal{C}_{\text{test}}\) are identified as novel categories.
The set of novel categories is expressed as \(\mathcal{C}_{\text{N}} = \mathcal{C} \setminus \mathcal{C}_{\text{B}} \neq \varnothing\).
For each category \(c \in \mathcal{C}\), we utilize CLIP to encode its text embedding \(t_c \in \mathbb{R}^d\), and $\mathcal{T}_\textsc{cls} = \{t_c\}_{c=1}^C$ (\(C\) is the size of the category vocabulary).

%% file: sections/3_method/overview_of_fcdetr.tex
\begin{figure}[tb]
    \centering
    \includegraphics[width=\linewidth]{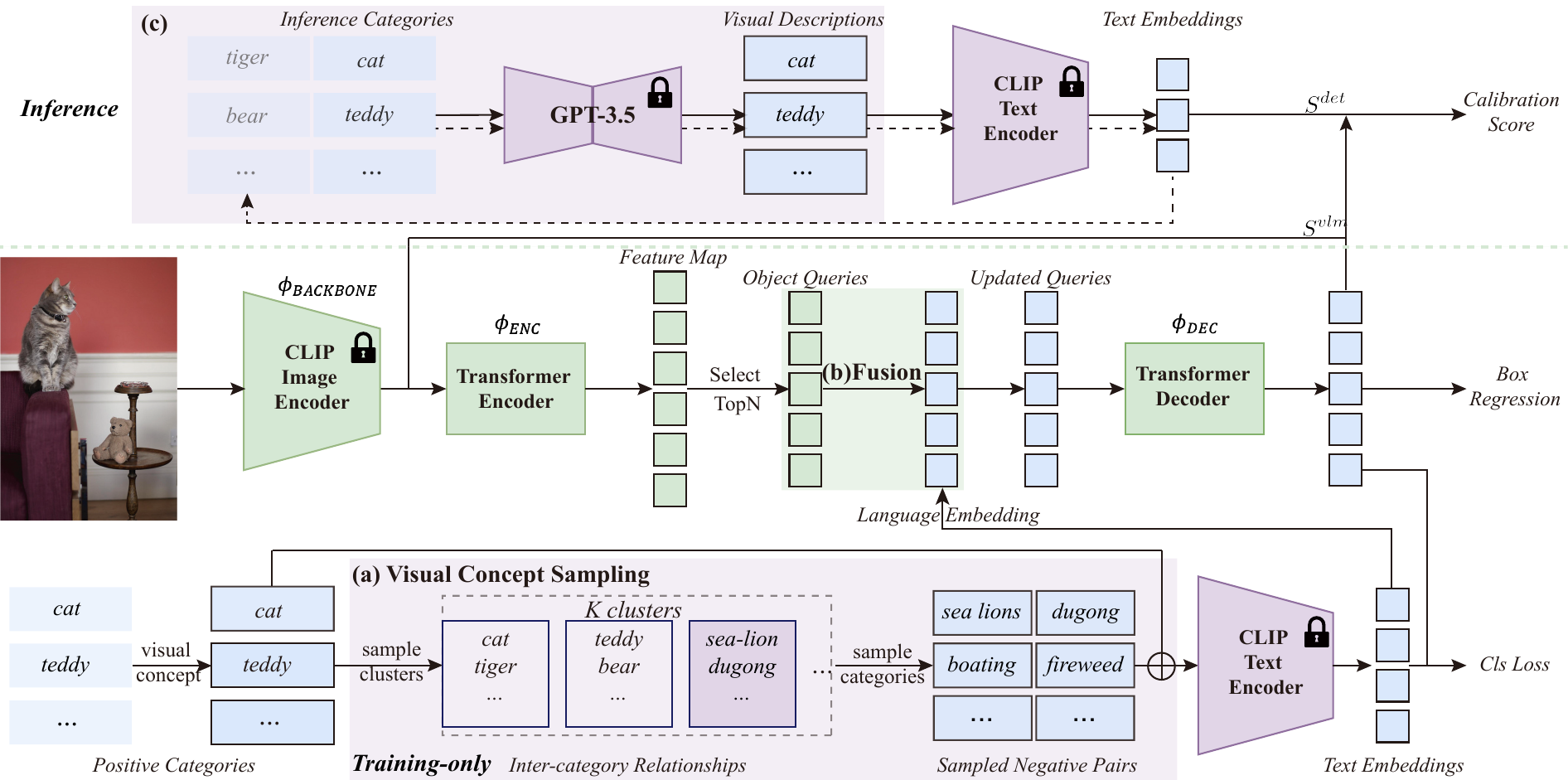}
    \caption{An overview of LaMI-DETR Framework.
    LaMI-DETR adapts the DETR model by incorporating a frozen CLIP image encoder as the backbone and replacing the final classification layer with CLIP text embeddings. 
    (a) Visual Concept Sampling, applied only during the training phase, leverages pre-extracted inter-category relationships to sample easy negative categories that are visually distinct from ground truth classes. This encourages the detector to derive more generalized foreground features rather than overfitting to base categories.
    (b) Language embeddings selected are integrated into the object queries for enhanced classification accuracy.
    (c) During inference, confusing categories are identified to improve VLM score.}
    \label{fig:framework}
\end{figure}

The overall framework of LaMI-DETR is illustrated in Figure~\ref{fig:framework}.
Given an image input, we obtain the spatial feature map using the ConvNext backbone from the pre-trained CLIP image encoder \(\left(\Phi_{\textsc{backbone}}\right)\), which remains frozen during training. 
Then the feature map is sequentially subjected to a series of operations:
a transformer encoder \(\left(\Phi_{\textsc{enc}}\right)\) to refine the feature map;
a transformer decoder \(\left(\Phi_{\textsc{dec}}\right)\), producing a set of query features \(\left\{f_j\right\}_{j=1}^{N}\);
The query features are then processed by a bounding box module \(\left(\Phi_{\textsc{bbox}}\right)\) to infer the positions of objects, denoted as \(\left\{\mathbf{b}_j\right\}_{j=1}^{N}\).
We follow the inference pipeline of F-VLM~\cite{fvlm} and use VLM score $S^{vlm}$ to calibrate detection score $S^{det}$.
\begin{align}
&\label{eqn:vlm_score}
S_j^{vlm} = \mathcal{T}_\textsc{cls}\cdot\Phi_\text{pooling}\left(b_j\right) \\
&\label{eqn:score}S_c^{cal} = 
    \begin{cases}
        {S^{vlm}_c}^\alpha \cdot {S^{det}_c}^{(1-\alpha)} & \text{if } c \in \mathcal{C}_B \\
        {S^{vlm}_c}^\beta \cdot {S^{det}_c}^{(1-\beta)} & \text{if } c \in \mathcal{C}_N
    \end{cases}
\end{align}
\subsubsection{\textbf{Comparison with other Open-Vocabulary DETR. }} 
CORA \cite{wu2023cora} and EdaDet \cite{Shi_2023_ICCV} also propose to use a frozen CLIP image encoder in DETR for extracting image features. However, LaMI-DETR significantly differs from these two approaches in the following aspects.

Firstly, regarding the number of backbones used, both LaMI-DETR and CORA employ a single backbone. In contrast, EdaDet utilizes two backbones: a learnable backbone and a frozen CLIP image encoder.

Secondly, both CORA and EdaDet adopt an architecture that decouples classification and regression tasks. While this method addresses the issue of failing to recall novel classes, it necessitates extra post-processing steps, such as NMS, disrupting DETR's original end-to-end structure.

Furthermore, both CORA and EdaDet require RoI-Align operations during training. In CORA, the DETR only predicts objectness, necessitating RoI-Align on the CLIP feature map during anchor pre-matching to determine the specific categories of proposals. EdaDet minimizes the cross-entropy loss based on each proposal's classification scores, obtained through a pooling operation. Consequently, CORA and EdaDet require multiple pooling operations during inference. In contrast, LaMI-DETR simplifies this process, needing only a single pooling operation at the inference stage.

%% file: sections/3_method/language_model_instruction.tex
Unlike previous methods that only rely on the vision-language alignment of VLMs, we aim to improve open-vocabulary detectors by enhancing concept representation and investigating inter-category relationships.
To achieve this, we first explain the process of constructing visual concepts and delineating their relationships. 
In \textit{Language Embedding Fusion} and \textit{Confusing Category} sections, we describe methods for more accurately representing concepts during the training and inference processes. 
The \textit{Visual Concept Sampling} section addresses how to mitigate overfitting issue through the use of inter-category relationships. 
Finally, we detail the distinctions with other research effort.

\subsubsection{\textbf{Inter-category Relationships Extraction. }}
Based on the problem identified in Figure~\ref{fig:cluster_c}, we employ visual descriptions to establish visual concepts, refining concept representation. Furthermore, we utilize T5, which possesses extensive textual semantic knowledge, to measure similarity relationships among visual concepts, thereby extracting inter-category relationships.

\begin{figure}[h]
    \centering
    \includegraphics[width=\linewidth]{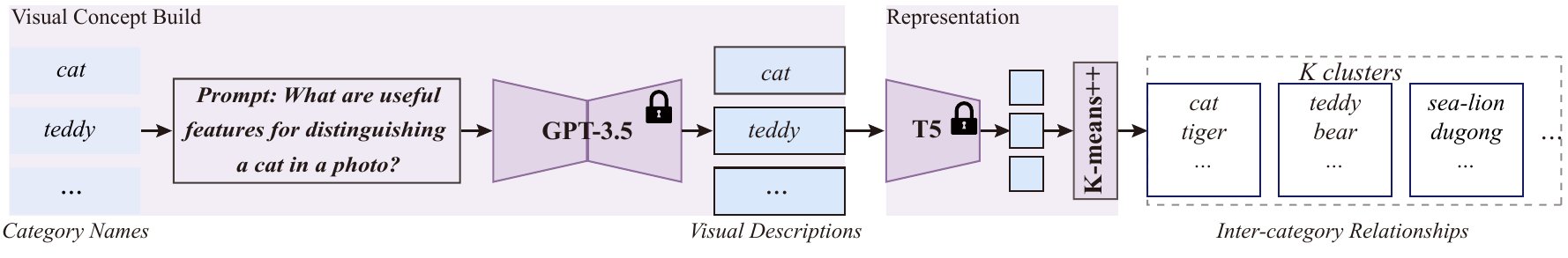}
    \caption{Illustration of Inter-category Relationships Extraction. 
    Visual descriptions generated by GPT-3.5 are processed by T5 to cluster categories with visual similarities. }
    \label{fig:framework_lami}
\end{figure}

As illustraed in Figure~\ref{fig:framework_lami}, given a category name \(c \in \mathcal{C}\), we extract its fine-grained visual feature descriptors \(d\) using the method described in~\cite{menon2022visual}. We define \(\mathcal{D}\) as the visual description space for categories in \(\mathcal{C}\). These visual descriptions \(d \in \mathcal{D}\) are then sent to the T5 model to obtain the visual description embeddings \(e \in \mathcal{E}\). Consequently, we construct an open set of visual concepts \(\mathcal{D}\) and their corresponding embeddings \(\mathcal{E}\).
To identify visually similar concepts, we propose clustering the visual description embeddings $\mathcal{E}$ into \(K\) cluster centroids. Concepts grouped under the same cluster centroid are deemed to possess similar visual characteristics. The extracted inter-category relationships are then applied in the visual concept sampling as shown in Figure~\ref{fig:framework}(a).

\subsubsection{\textbf{Language Embedding Fusion. }}
\label{chap:lef}
As shown in Figure~\ref{fig:framework}(b), after transformer encoder, each pixel on the feature map \(\{f_i\}_{i=1}^{M}\) is interpreted as an object query, with each directly predicting a bounding box. To select the top \(N\) scoring bounding boxes as region proposals, the process can be encapsulated as follows:
\begin{align}
\{q_j\}_{j=1}^{N} = \text{Top}_N(\{\mathcal{T}_\textsc{cls} \cdot f_i\}_{i=1}^M). 
\end{align}
In LaMI-DETR, we fuse each query \(\{q_j\}_{j=1}^{N}\) with its closest text embedding, resulting in:
\begin{align}
\label{eqn:embedding_fusion}
\{q_j\}_{j=1}^{N} = \{q_j \oplus t_j\}_{j=1}^{N},
\end{align}
where \(\oplus\) denotes element-wise addition.

On one hand, the visual descriptions are sent to the T5 model to cluster visually similar categories, as previously described. On the other hand, the visual descriptions \(d_j \in \mathcal{D}\) are forwarded to the text encoder of the CLIP model to update the classification weights, denoted as \(\mathcal{T}_\textsc{cls} = \{t'_c\}_{c=1}^{C}\), where \(t'_c\) represents the text embedding of $d$ in the CLIP text encoder space. Consequently, the text embeddings used in the language embedding fusion process are updated accordingly:
\begin{align}
\label{eqn:embedding_update}
\{q_j\}_{j=1}^{N} = \{q_j \oplus t'_j\}_{j=1}^{N}
\end{align}

\subsubsection{\textbf{Confusing Category. }}
Due to similar visual concepts often sharing common features, nearly identical visual descriptors can be generated for these categories. 
This similarity poses challenges in distinguishing similar visual concepts during the inference process.

To distinguish easily confusable categories during the inference process, we initially identify the most similar category \(c^{\text{conf}} \in \mathcal{C}\) for each class \(c \in \mathcal{C}\) within the CLIP text encoder semantic space, based on \(\mathcal{T}_\textsc{cls}\). We then modify the prompt for generating visual descriptions \(d' \in \mathcal{D}'\) for category \(c\) to emphasize the features that differentiate \(c\) from \(c^{\text{conf}}\). Let \(t''\) be the text embedding of \(d'\) in the CLIP text encoder space. As shown in Figure~\ref{fig:framework}(c), We update the inference pipeline as follows:
\begin{align}
\label{eqn:confusing_category}
& \mathcal{T}'_\textsc{cls} = \{t''_c\}_{c=1}^C, \\
& S_j^{vlm} = \mathcal{T}'_\textsc{cls} \cdot \Phi_{\text{pooling}}\left(b_j\right).
\end{align}
\subsubsection{\textbf{Visual Concept Sampling. }}
To address the challenges posed by incomplete annotations in open-vocabulary detection datasets, we employ Federated Loss~\cite{zhou2021probablistic}, originally introduced for long-tail datasets~\cite{gupta20lvis}. This approach involves randomly selecting a set of categories to calculate detection losses for each minibatch, effectively minimizing issues related to missing annotations in certain classes. 
Given category occurrence frequency \(p = [p_1, p_2, \ldots, p_C]\), where \(p_c\) denotes the occurrence frequency in training data of the \(c^{\text{th}}\) visual concept and \(C\) represents the total number of categories. 
We randomly draw \(C_{\text{fed}}\) samples based on the probability distribution \(p\). The likelihood of selecting the \(c^{\text{th}}\) sample \(x_c\) is proportional to its corresponding weight \(p_c\). 
This method facilitates the transfer of visual similarity knowledge, extracted by the language model, to the detector, thereby reducing the issue of overfitting:
\begin{align}
\label{eqn:federated_loss}
P(X = c) = p_c, \quad \text{for } c = 1, 2, \ldots, C
\end{align}
Incorporating federated loss, the classification weight is reformulated as \(\mathcal{T}_\textsc{cls} = \{t''_c\}_{c=1}^{C_{\text{fed}}}\), where \(\mathcal{C}_\textsc{fed}\) denotes the categories engaged in the loss calculation of each iteration, and \(C_{\text{fed}}\) is the count of \(\mathcal{C}_\textsc{fed}\).

We utilize a frozen CLIP with strong open vocabulary capabilities as LaMI-DETR's backbone. However, due to the limited categories in detection datasets, overfitting to base classes is inevitable after training.
To mitigate overtraining on base categories, we aim to sample straightforward negative categories based on the results of visual concepts clustering. 
In LaMI-DETR, let the clusters containing the ground truth categories be denoted by \(\mathcal{K}_G\) in a given iteration. We denote all the categories within \(\mathcal{K}_G\) as \(\mathcal{C}_g\). 
Specifically, we aim to exclude \(\mathcal{C}_g\) from being sampled in the current iteration. To achieve this, we set the frequency of occurrence for categories within \(\mathcal{C}_g\) to zero.
This approach enables the transfer of visual similarity knowledge, extracted by the language model, to the detector, mitigating overfitting issue:
\begin{align}
\label{eqn:concept_sampling}
&p_c^{cal} = 
    \begin{cases}
        0 & \text{if } c \in \mathcal{C}_g \\
        p_c & \text{if } c \notin \mathcal{C}_g
    \end{cases}
\end{align}
where \(p_c^{cal}\) indicates the frequency of occurrence of category \(c\) after language model calibration, ensuring visually similar categories are not sampled during this iteration. This process is shown in Figure~\ref{fig:framework}(a).

\subsubsection{\textbf{Comparison with concept enrichment. }}
The visual concept description is different from the concept enrichment employed in DetCLIP~\cite{yao2022detclip}. 
The visual descriptions used in LaMI place more emphasis on the visual attributes inherent to the object itself. 
In DetCLIP, category label is supplemented with definitions, which may include concepts not present in the pictures to rigorously characterize a class. 

%% file: sections/4_exp/overview.tex
\section{Experiments}

Section~\ref{sec:dataset} introduces the standard dataset and benchmarks commonly utilized in the field, as detailed in~\cite{vild2021}. 
Section~\ref{sec:implementation} outlines the implementation and training details of our LaMI-DETR, which leverages knowledge of visual characteristics from language models. 
We present a comparison of our models with existing works in Section~\ref{sec:main}, showcasing state-of-the-art performance. Additionally, Section~\ref{sec:main} includes results on cross-dataset transfer to demonstrate the generalizability of our approach. Finally, Section~\ref{sec:ablation} conducts ablation studies to examine the impact of our design decisions.

\subsection{Datasets}
\label{sec:dataset}
\input{sections/4_exp/dataset}

\subsection{Implementation Details}
\label{sec:implementation}
\input{sections/4_exp/implement}

\subsection{Open-Vocabulary Detection Results}
\label{sec:main}
\input{sections/4_exp/benchmark}

\subsection{Ablation Study}
\label{sec:ablation}
\input{sections/4_exp/ablation}

%% file: sections/4_exp/dataset.tex
\subsubsection{\textbf{LVIS.}}
Our experiments are conducted on the LVIS dataset, which includes annotations for \(1,203\) object categories. 
These categories are divided into three groups—rare, common, and frequent—based on the number of training images containing a given class. 
Following the approach of previous studies, we categorize them into \(866\) base classes, encompassing frequent and common categories, and \(337\) novel classes, consisting of rare categories. 
To create an open-vocabulary scenario, we exclude annotations for novel classes from the training images. 
In line with standard practice, we report the mean average precision (mAP) for predicted boxes specifically for the rare classes, denoted as AP\(_\textrm{r}\). Additionally, we present the box AP averaged across all classes to reflect overall performance, denoted as mAP.

\subsubsection{\textbf{Object365 and VisualGenome.}}
For a fair comparison with OWL-ViT~\cite{owlvit2023, owlvitv22023scaling}, we adopt the same training settings, utilizing data from Object365 and VisualGenome. 
To conserve training time, we employ only a \(1/3\) random sample of Object365 in our study. 
With respect to VisualGenome, we meticulously replicate OWL-ViT's preprocessing steps by eliminating all detection annotations that correspond to the names of LVIS's rare categories. The resulting curated dataset is referred to as VG dedup.

%% file: sections/4_exp/implement.tex
Training is conducted on \(8\) 40G A100 GPUs with a total batch size of \(32\). 
For the OV-LVIS setting, we train the model for \(12\) epochs. 
In the VG-dedup benchmark, to ensure a fair comparison with OWL-ViT, we initially pretrain LaMI-DETR on a randomly sampled \(1/3\) of the Object365 dataset for \(12\) epochs. Subsequently, LaMI-DETR is finetuned on the VG dedup dataset for an additional \(12\) epochs.

The detector utilizes ConVNext-Large~\cite{liu2022convnet} from OpenCLIP~\cite{openclip} as its backbone, which remains frozen throughout the training process. 
LaMI-DETR, building upon DINO, employs \(900\) queries as specified in detrex~\cite{ren2023detrex}. 
We adhere closely to the original training configurations detailed in detrex, with the exception of employing an exponential moving average (EMA) strategy to enhance training stability. 
To balance the distribution of training samples, we apply repeat factor sampling~\cite{gupta20lvis} using the default hyperparameters. 
For federated loss, the numbers of categories \(C_{\text{fed}}\) are set to \(100\) and \(700\) for OV-LVIS and VG dedup datasets, respectively. 

To explore a broader range of visual concepts for more effective clustering, we compile a comprehensive category collection from LVIS, Object365, VisualGenome, Open Images, and ImageNet-21K. 
Redundant concepts are filtered out using WordNet hypernyms, resulting in a visual concept dictionary comprising \(26,410\) unique concepts. 
During the visual concept grouping phase, this dictionary is clustered into \(K\) centers, with \(K\) being \(128\) for OV-LVIS and \(256\) for VG dedup, respectively.

%% file: sections/4_exp/benchmark.tex
\subsubsection{\textbf{OV-LVIS.}}

\input{tables/ov_lvis}
We compare our LaMI-DETR framework with the other state-of-the-art OVOD methods in Table~\ref{tab:ov_lvis}.
We report overall box AP performance and box AP for "rare" classes only.
The latter metric is the key measure of OVOD performance.
Our method obtain the best performance on both AP$_\textrm{r}$ and overall mAP compared to existing approaches for open-vocabulary object detection, while utilizing a more challenging strictly open-vocabulary training paradigm without additional data.
LaMI-DETR, with a backbone of only 196M parameters significantly less than CFM-ViT's 303M achieves superior performance. 
Moreover, LaMI-DETR does not utilize additional image-level datasets.
The results demonstrate that LaMI-DETR has lower computational requirements and higher accuracy.

\subsubsection{\textbf{Zero-shot LVIS.}}
\input{tables/vg_dedup}
We evaluate the model's ability to recognize diverse and rare objects on LVIS in a zero-shot setting. We replace VG-dedup with LVIS vocabulary embeddings for zero-shot detection without finetuning. 
We assume all categories are novel and set $\alpha, \beta$=(0.0, 0.25) in Eq~\ref{eqn:score}.
We use OWL as the baseline for our models. 
The results are shown in Table~\ref{tab:vg_degup}. LaMI-DETR outperforms OWLs under the same settings.

\subsubsection{\textbf{Cross-dataset Transfer.}}
To evaluate the generalizability of our method in a cross-dataset transfer detection setting, we conduct experiments on the COCO and Objects365-v1 validation split. Specifically, we directly applies the detector trained on the LVIS base categories, while replacing the LVIS class embeddings with those of COCO/Objects365 for transfer detection without further finetuning. 
All categories were treated as novel.
Our best-performing model achieved 42.8 AP on COCO and 21.9 AP on Object365, outperforming CoDet by +3.7 AP on COCO and CFM by +3.2 AP on Object365 according to Table~\ref{tab:transfer}. 
\input{tables/transfer}

%% file: tables/ov_lvis.tex
\begin{table}[t]
    \centering
    \caption{
    \small\textbf{LVIS open-vocabulary detection (box AP).} LaMI-DETR outperforms the best existing approach by +7.8 box AP$_r$ in the standard benchmark. All methods use the same instance-level supervision from LVIS base categories for detection training. $\dagger$: reports mask AP. $\star$: uses the image-level data in pretraining. 
    We calculate the backbone's parameters based on models released by CLIP except RN50, which may vary slightly from their actual sizes.}
    {
    \begin{tabular}{lccccll}
        \toprule
        Method & \makecell{Pretrained \\Model} & \makecell{Detector \\Backbone} & \makecell{Backbone \\Size} & \makecell{Image-level \\Dataset} & \bf{AP$_\textrm{r}$} & \gray{AP}
        \\
        \midrule
        VL-PLM~\cite{vlplm2022}                & ViT-B/32  & R-50      &26M  &IN-L   & 17.2\textsuperscript{$\dagger$}      & \gray{27.0}\textsuperscript{$\dagger$} \\
        OV-DETR~\cite{zang2022open}            & ViT-B/32  & R-50      &26M  &\xmark & 17.4\textsuperscript{$\dagger$}      & \gray{26.6}\textsuperscript{$\dagger$} \\
        DetPro-Cascade~\cite{detpro}           & ViT-B/32  & R-50      &26M  &\xmark & 21.7 & \gray{30.5} \\
        Rasheed~\cite{Hanoona2022Bridging}     & ViT-B/32  & R-50      &26M  &IN-L   & 21.1\textsuperscript{$\dagger$}      & \gray{25.9}\textsuperscript{$\dagger$} \\
        PromptDet~\cite{feng2022promptdet}     & ViT-B/32  & R-50      &26M  &LAION-novel       & 21.4\textsuperscript{$\dagger$}      & \gray{25.3}\textsuperscript{$\dagger$} \\
        OADP~\cite{OADP}                       & ViT-B/32  & R-50      &26M  &\xmark       & {21.9}  & \gray{28.7} \\
        RegionCLIP~\cite{regionclip2022}       & R-50x4    & R-50x4    &87M  &CC3M       & 22.0\textsuperscript{$\dagger$}      & \gray{32.3}\textsuperscript{$\dagger$}\\
        CORA~\cite{wu2023cora}                 & R-50x4    & R-50x4    &87M  &\xmark       &22.2 &- \\
        BARON~\cite{wu2023baron}               & ViT-B/32  & R-50      &26M  &CC3M       & 23.2 & \gray{29.5} \\
        CondHead~\cite{condhead}               & R-50x4    & R-50x4    &87M  &CC3M       & 25.1 & \gray{33.7} \\
        Detic-CN2~\cite{detic2022}             & ViT-B/32  & R-50      &26M  &IN-L       & 24.6\textsuperscript{$\dagger$}      & \gray{32.4}\textsuperscript{$\dagger$} \\
        ViLD-Ens~\cite{vild2021}               & ViT-B/32  & R-50      &26M  &\xmark       &16.7 & \gray{27.8}  \\
        F-VLM~\cite{fvlm}                      & R-50x64   & R-50x64   &420M &\xmark       & 32.8\textsuperscript{$\dagger$}      & \gray{34.9}\textsuperscript{$\dagger$} \\
        OWL-ViT~\cite{owlvit2023}              & ViT-L/14  & ViT-L/14  &306M  &\xmark       & 25.6     & \gray{34.7} \\
        {RO-ViT~\cite{rovit2023}}              & ViT-B/16  & ViT-B/16  &86M  &ALIGN\textsuperscript{$\star$}       & {28.4} & \gray{31.9}\\
        {RO-ViT~\cite{rovit2023}}              & ViT-L/16  & ViT-L/16  &303M &ALIGN\textsuperscript{$\star$}       &{33.6}& \gray{36.2} \\
        CFM-ViT~\cite{kim2023contrastivecfm}   & ViT-B/16  & ViT-B/16  &86M  &ALIGN\textsuperscript{$\star$}       & {29.6}& \gray{33.8} \\
        CFM-ViT~\cite{kim2023contrastivecfm}   & ViT-L/16  & ViT-L/16  &303M &ALIGN\textsuperscript{$\star$}       & {35.6}& \gray{38.5} \\
        \bf{ours}             & ConVNext-L  & ConVNext-L  &196M  &\xmark       & \bf{43.4}& \bf{\gray{41.3}} \\
        \bottomrule
    \end{tabular}
    }
    \label{tab:ov_lvis}
\end{table}

%% file: tables/vg_dedup.tex
\begin{table}[t]
    \centering
    \caption{
    \small\textbf{LVIS zero-shot detection (box AP).} \S: The models only report fixed AP~\cite{fixedap} on \emph{LVIS-val}. The models depicted in this figure utilize multiple detection datasets, excluding LVIS; therefore, we refer to this configuration as the zero-shot setting.
    }
    {\footnotesize
    \begin{tabular}{lcccll}
        \toprule
        Method & \makecell{Detector \\Backbone} & \makecell{Datasets} & \bf{AP$_\textrm{r}$} & \gray{AP}
        \\
        \midrule
        GLIP-L~\cite{li2021glip}            & Swin-L       &O365,GoldG,Cap4M       & 17.1     & \gray{26.9}  \\
        GroundingDINO~\cite{groundingdino}  & Swin-L       &O365,GoldG,OI,Cap4M,COCO,RefC  & 22.0     & \gray{32.3}  \\
        DetCLIP\textsuperscript{$\S$}~\cite{yao2022detclip}       & Swin-L       &O365,GoldG,YFCC1M      & \gray{27.6}     & \gray{31.2}  \\
        DetCLIPv2\textsuperscript{$\S$}~\cite{yao2023detclipv2}   & Swin-L       &O365,GoldG,CC15M       & \gray{33.3}     & \gray{36.6}  \\
        OWL-ViT~\cite{owlvit2023}           & ViT-L/14     &O365,VG-dedup          & 31.2     & \gray{34.6}  \\
        OWL-ST~\cite{owlvitv22023scaling}   & ViT-L/14     &O365,VG-dedup          & 34.9     & \gray{33.5} \\
        \bf{ours}                            & ConVNext-L   &O365,VG-dedup          & \bf{37.8}   & \gray{35.4} \\
        \bottomrule
    \end{tabular}
    }
    \label{tab:vg_degup}
\end{table}

%% file: tables/transfer.tex
\begin{table}[]
    \centering
    \caption{
        Cross-datasets transfer detection from OV-LVIS to COCO and Objects365. F-VLM adopts RN50 in CLIP as backbone, which is larger than standard RN50.
    }
    \begin{tabular}{lcccccccc}
    \toprule
    \multirow{2}{*}{Method} & \multirow{2}{*}{Backbone} & \multirow{2}{*}{Parameters} & \multicolumn{3}{c}{COCO} & \multicolumn{3}{c}{Objects365} \\
    \cmidrule(lr){4-6} \cmidrule(lr){7-9}
    & & & AP & AP$_{50}$ & AP$_{75}$ & AP & AP$_{50}$ & AP$_{75}$ \\
    \midrule
    ViLD~\cite{vild2021}   &RN50 &26M & 36.6 & 55.6 & 39.8 & 11.8 & 18.2 & 12.6 \\
    DetPro~\cite{detpro} &RN50 &26M & 34.9 & 53.8 & 37.4 & 12.1 & 18.8 & 12.9 \\
    F-VLM~\cite{fvlm}  &RN50 &38M & 32.5 & 53.1 & 34.6 & 11.9 & 19.2 & 12.6 \\
    BARON~\cite{wu2023baron}  &RN50 &26M & 36.2 & 55.7 & 39.1 & 13.6 & 21.0 & 14.5 \\
    CoDet~\cite{ma2023codet} &EVA02-L &304M & 39.1 & 57.0 & 42.3 & 14.2 & 20.5 & 15.3 \\
    CFM~\cite{kim2023contrastivecfm} &ViT-L/16 &303M & - & - & - & 18.7 & 28.9 & 20.3 \\
    \bf{ours} &ConvNext-L &196M & \textbf{42.8} & \textbf{57.6} & \textbf{46.9} & \textbf{21.9} & \textbf{30.0} & \textbf{23.5} \\
    
    \bottomrule
    \end{tabular}
\label{tab:transfer}
\end{table}

%% file: sections/4_exp/ablation.tex
To study the advantages of LaMI-DETR, we provide ablation studies on the OV-LVIS benchmark. 

\subsubsection{\textbf{LaMI-DETR.}}
\input{tables/ablation_lamidetr}
Table~\ref{tab:ablation_lamidetr} demonstrates the impact of incorporating language model guidance into our LaMI-DETR framework. 
The version without LaMI module achieves an AP\(_\textrm{r}\) of \(33.0\). By integrating our proposed LaMI module, the model achieves an AP\(_\textrm{r}\) of \(43.4\).
The top two rows in Table~\ref{tab:ablation_lamidetr} shows language embedding fusion in Eq.\ref{eqn:embedding_fusion} brings a 0.8 AP\(_\textrm{r}\) gain.
The $3$\textsuperscript{nd} to  $5$\textsuperscript{th} row in Table~\ref{tab:ablation_lamidetr} adds Visual Concept Sampling, embedding update and Confusing Category distinguishing to baseline gradually. 

\subsubsection{\textbf{Confusing Category.}}
\input{tables/ablation_confuse}
We demonstrate the effectiveness of Confusing Category in Table~\ref{tab:ablation_confuse}.
Given the ground truth bounding boxes, we use different text embeddings to classify their region features. 
To evaluate the performance, we compute "Mean Accuracy" (accuracy for each category independently with equal weights).
For the following strategies, we use RoI-Align to directly extract features from CLIP.
The table validates that the CLIP text encoder can discriminate categories from confusing ones with our refined concept representation.

\subsubsection{\textbf{The Cluster Design.}}
\input{tables/ablation_cluster}
Visual Concept Sampling aims to sample negative categories with large visual differences from the positive class, enabling the detector to utilize inter-class relationships by penalizing categories with large visual differences, thus achieving generalization to visually close classes. 
We validate this claim through the enhancements detailed in Table~\ref{tab:ablation_sample}. 

Our experimental results in Table \ref{tab:ablation_sample} demonstrate the effectiveness of our sampling negtive classes method.
The first row shows results of baseline. 
Rows $2$\textsuperscript{nd}-$5$\textsuperscript{th} employ the Visual Concept Sampling module but vary the clustering method.
Specifically, the second row clusters category name embeddings from a CLIP text encoder, corresponding to (a) in Figure~\ref{fig:cluster_c}. 
The third row clusters category name embeddings from the T5 space, corresponding to Figure~\ref{fig:cluster_c}(b). 
The fourth row aims to match DetCLIP's concept enrichment by clustering definition embeddings in the T5 space. 
Finally, the last row presents our full method, clustering category visual description embeddings from the T5 space as Figure~\ref{fig:cluster_c}(c). 
This systematic ablation analyzes how different semantics and grouping strategies within the Visual Concept Sampling module affect downstream detection performance, validating the importance of visual similarity-based concept sampling for our task.

%% file: tables/ablation_lamidetr.tex
\begin{table}[t]
    \centering
    \caption{
    Ablations for our model. Language Model Instruction consists of visual concepts sampling, embedding update and confusing categories distinguish.
    Below the horizontal line are the results with the class factor. See Table~\ref{tab:ablation_revise} for details.
    }
    \begin{tabular}{lcccccc}
    \toprule
    \rowNumber{\#} & 
    \makecell{Federated\\Loss(Eq.\ref{eqn:federated_loss})} &
    \makecell{Embedding\\Fusion(Eq.\ref{eqn:embedding_fusion})} & 
    \makecell{\footnotesize Visual Concepts\\Sampling(Eq.\ref{eqn:concept_sampling})} &  
    \makecell{\footnotesize Embedding\\Update(Eq.\ref{eqn:embedding_update})}& 
    \makecell{\footnotesize Confusing \\Category(Eq.\ref{eqn:confusing_category})} & 
    AP$_\textrm{r}$ \\
    \cmidrule(r){1-2}
    \cmidrule(r){3-3}
    \cmidrule(r){4-4}
    \cmidrule(r){5-5}
    \cmidrule(r){6-6}
    \cmidrule(r){7-7}
    \rowNumber{1} &\cmark & & & & &32.2 \\
    \rowNumber{2} &\cmark &\cmark & & & &33.0 \\
    \midrule
    \rowNumber{3} &\cmark &\cmark &\cmark & & &40.1 \\
    \rowNumber{4} &\cmark &\cmark &\cmark &\cmark & &42.5 \\
    \rowNumber{5} &\cmark &\cmark &\cmark &\cmark &\cmark &43.4 \\
    \bottomrule
    \end{tabular}
    \label{tab:ablation_lamidetr}
\end{table}

%% file: tables/ablation_confuse.tex
\begin{table}[]
\centering
\caption{Ablation study on the confusing category. 
Zero-shot proposal classification performance on LVIS minival datasets.
}
\begin{tabular}{@{}ccccccc@{}}
\toprule
Model & mAcc$_\textrm{r}$ & mAcc$_\textrm{c}$ & mAcc$_\textrm{f}$ & mAcc\\
    \midrule
    CLIP & 43.8 & 44.1 & 37.8 & 41.0\\
    visual desc.~\cite{menon2022visual} & 49.5 & 45.8 & 40.2 & 43.4 \\
    ours & 52.7 & 46.1 & 41.4 & 44.4 \\
\bottomrule
\end{tabular}
\label{tab:ablation_confuse}

\end{table}

%% file: tables/ablation_cluster.tex
\begin{table}[]
\centering
\caption{Ablation study on the cluster designs. For fair comparison, all detectors use classification weights from CLIP text encoder name embeddings. $\dagger$: Results with class factor. See Table~\ref{tab:ablation_revise} for details.}
\begin{tabular}{@{}ccccccc@{}}
\toprule
Model & Cluster Encoder & Cluster Text & AP$_\textrm{r}$ & AR$_\textrm{r}$\\
    \midrule
    baseline & - & - & 33.0 & 40.3\\
    baseline+VCS & CLIP Text Encoder & name & 33.5 & 41.4 \\
    baseline+VCS & Instructor Embedding & name & 34.1 & 39.5 \\
    baseline+VCS & Instructor Embedding & name+definition & 31.5 & 37.3\\
    baseline+VCS & Instructor Embedding & name+visual desc. & 40.1\textsuperscript{$\dagger$} & 57.0\textsuperscript{$\dagger$}\\
\bottomrule
\end{tabular}
\label{tab:ablation_sample}

\end{table}

%% file: sections/5_conclusion.tex
\section{Conclusion}
In this paper, we undertake the first effort to explore inter-category relationships for generalization in OVOD. 
We introduce LaMI-DETR, a framework that effectively utilize the visual concepts similarity to sample negtive categories during training for learning generalizable object localization and retaining open vocabulary knowledge of VLMs.
Additionally, the refined concepts enable effective object classification especially between confusing categoris. 
Experiments show that LaMI-DETR achieves state-of-the-art performance across various OVOD benchmarks.
On the other hand, our method utilizes the CLIP ConvNext-L architecture as the visual backbone. Exploring alternative pre-trained VLMs such as those based on ViT is under-explored here. We leave this for further investigation.

%% file: sections/6_supp.tex
\section{Supplementary Material}

\subsection{Visualization}
\begin{figure}[h]
    \centering
    \includegraphics[width=\linewidth]{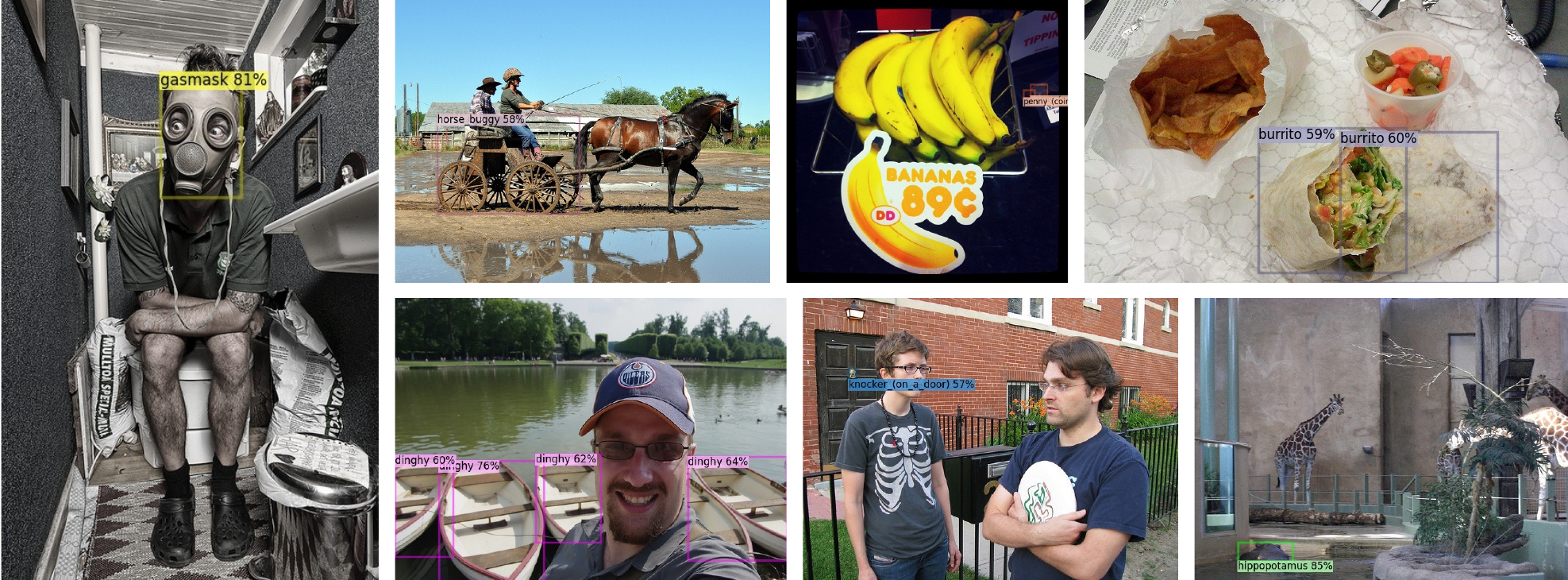}
    \caption{Visualization of results by LaMI-DETR on OV-LVIS. For better clarity, we only display the prediction results for novel categories.
    }
    \label{fig:lvisnovel}
\end{figure}

We visualize detection results of LaMI-DETR on LVIS novel categories (Figure~\ref{fig:lvisnovel}). 

\subsection{Ablation}
In the OVD setting, there exist both base and novel categories during inference. The logits for novel classes are usually lower than those for base categories. This issue is commonly alleviated by rescoring novel categories~\cite{wu2023cora}.
We multiply the logit of novel classes by a factor of 5.0 during inference.
We include results related to the factor in Table~\ref{tab:ablation_revise}.
\input{tables/ablation_revise}

\subsection{Further Analysis on generalization of LaMI}
\begin{figure}[h]
    \centering
    \includegraphics[width=\linewidth]{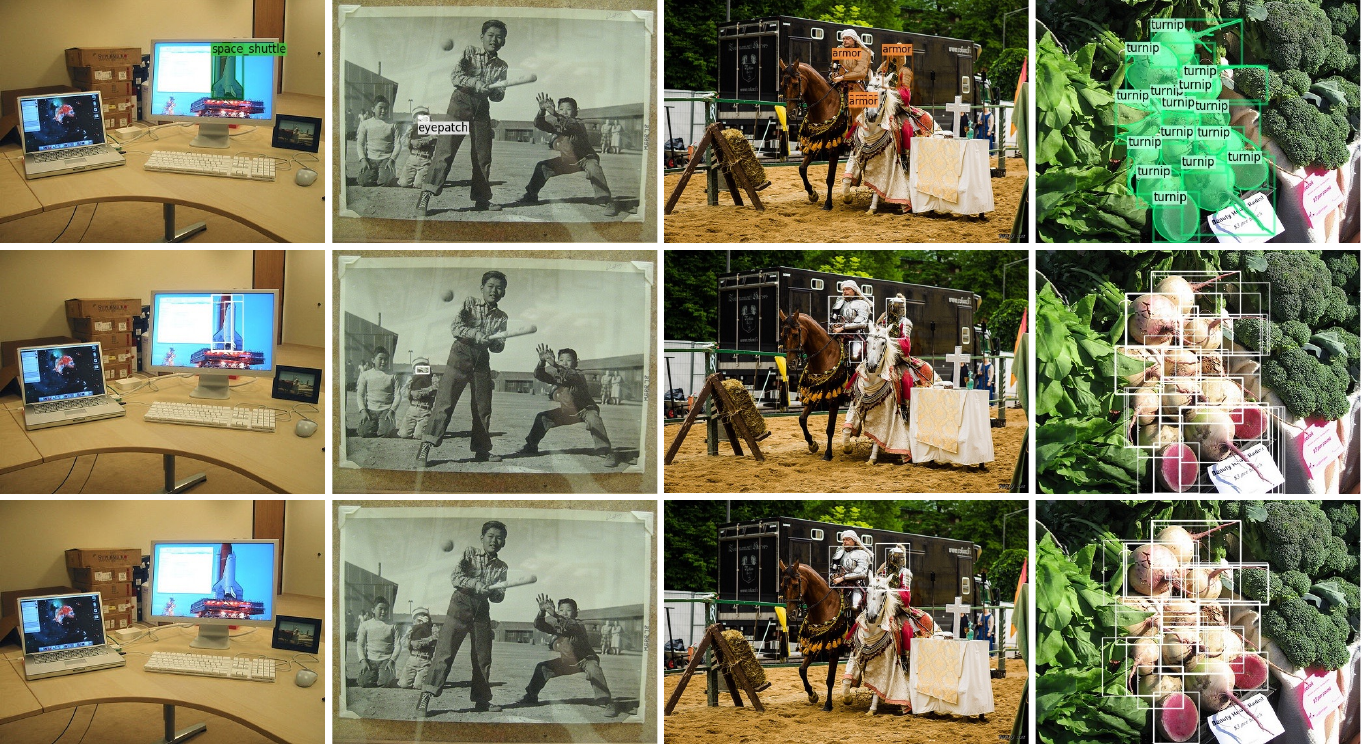}
    \caption{
    Visualization of proposals generated by the model with and without LaMI. Sequentially from top to bottom, each row displays the results for the ground-truth, LaMI-DETR, and the baseline, respectively. For detailed examination, please zoom in.
    }
    \label{fig:proposal}
\end{figure}

Figure~\ref{fig:proposal} illustrates the base-to-novel generalization capability of LaMI. Specifically, it employs models trained on the OV-LVIS benchmark to generate proposals. We visualize proposals having an IoU > 0.5 with the nearest ground-truth box for novel categories in the LVIS validation set.

\subsection{Confusing Category Details}
We provide a detailed description of the Confusing Category module pipeline in LaMI. Based on text embeddings from the CLIP text encoder, we identify visually similar categories for each inference category. 
Our method then constructs tailored prompts for GPT by incorporating disambiguating context about the confusable categories.
\begin{figure}[h]
    \centering
    \includegraphics[width=\linewidth]{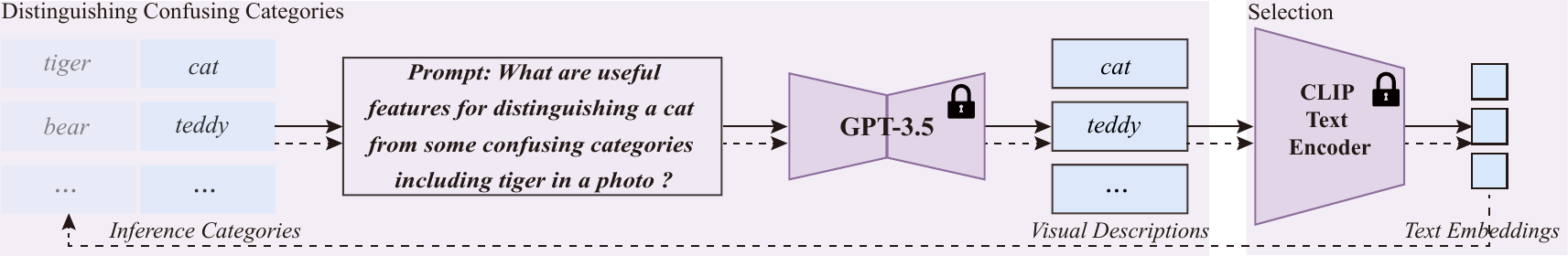}
    \caption{Illustration of Confusing Category module. 
    }
    \label{fig:visual_desc}
\end{figure}

\subsection{Inference Time}
\input{tables/fps_benchmark}
During inference, confusing categories are first selected using cosine similarity with sklearn. Next, API calls regenerate descriptions, followed by updating classifier weights. Finally, the model runs at 4.5 FPS.
We report FPS reflecting wall-clock time in tab~\ref{tab:fps}.

%% file: tables/ablation_revise.tex
\begin{table}[]
\centering
\caption{Novel classes factor. $\dagger$: results with factor.}
\begin{tabular}{@{}ccccccc@{}}
\toprule
Model & Cluster Encoder & Cluster Text & AP$_\textrm{r}$ & \gray{AP}\\
    \midrule
    baseline & - & - & 33.0 & \gray{40.6}\\
    baseline+VCS & Instructor Embedding & name+visual desc. & 34.2 & \gray{41.7}\\
    baseline+VCS\textsuperscript{$\dagger$} & Instructor Embedding & name+visual desc. & 40.1 & \gray{40.5}\\
    baseline+LaMI & Instructor Embedding & name+visual desc. & 41.7 & \gray{41.1}\\
    baseline+LaMI\textsuperscript{$\dagger$} & Instructor Embedding & name+visual desc. & 43.4 & \gray{41.3}\\
\bottomrule
\end{tabular}
\label{tab:ablation_revise}

\end{table}

%% file: tables/fps_benchmark.tex
\begin{table}[]
    \centering
    \caption{Zero-shot Evaluation on LVIS-minival. The FPS is evaluated on NVIDIA V100 GPU. To highlight our model's efficiency, we compare with methods using lighter backbones like Swin-T.}
    \begin{tabular}{llcl c cccc}
    \toprule
        Method & Backbone & FPS$\uparrow$ \\
        \hline
        GLIP-T &Swin-T & 0.12  \\
        GLIPv2-T & Swin-T  & 0.12 \\
        Grounding DINO-T & Swin-T& 1.5  \\
        DetCLIP-T & Swin-T & 2.3  \\
        LaMI-DETR & ConvNext-L & \textbf{4.5}  \\
    \bottomrule
    \end{tabular}
    \label{tab:fps}
\end{table}

%% file: main.bbl
\begin{thebibliography}{10}
\providecommand{\url}[1]{\texttt{#1}}
\providecommand{\urlprefix}{URL }
\providecommand{\doi}[1]{https://doi.org/#1}

\bibitem{bansal2018zero}
Bansal, A., Sikka, K., Sharma, G., Chellappa, R., Divakaran, A.: Zero-shot object detection. In: ECCV (2018)

\bibitem{brown2020languagegpt3}
Brown, T., Mann, B., Ryder, N., Subbiah, M., Kaplan, J.D., Dhariwal, P., Neelakantan, A., Shyam, P., Sastry, G., Askell, A., et~al.: Language models are few-shot learners. NIPS  \textbf{33},  1877--1901 (2020)

\bibitem{chen2024lw}
Chen, Q., Su, X., Zhang, X., Wang, J., Chen, J., Shen, Y., Han, C., Chen, Z., Xu, W., Li, F., et~al.: Lw-detr: A transformer replacement to yolo for real-time detection. arXiv preprint arXiv:2406.03459  (2024)

\bibitem{fixedap}
Dave, A., Doll{\'a}r, P., Ramanan, D., Kirillov, A., Girshick, R.: Evaluating large-vocabulary object detectors: The devil is in the details. arXiv preprint arXiv:2102.01066  (2021)

\bibitem{demirel2018zero}
Demirel, B., Cinbis, R.G., Ikizler-Cinbis, N.: Zero-shot object detection by hybrid region embedding. In: BMVC (2018)

\bibitem{detpro}
Du, Y., Wei, F., Zhang, Z., Shi, M., Gao, Y., Li, G.C.: Learning to prompt for open-vocabulary object detection with vision-language model. CVPR  (2022)

\bibitem{feng2022promptdet}
Feng, C., Zhong, Y., Jie, Z., Chu, X., Ren, H., Wei, X., Xie, W., Ma, L.: Promptdet: Towards open-vocabulary detection using uncurated images. In: ECCV. pp. 701--717. Springer (2022)

\bibitem{goodfellow2014generative}
Goodfellow, I., Pouget-Abadie, J., Mirza, M., Xu, B., Warde-Farley, D., Ozair, S., Courville, A., Bengio, Y.: Generative adversarial nets. In: NIPS (2014)

\bibitem{vild2021}
Gu, X., Lin, T.Y., Kuo, W., Cui, Y.: Open-vocabulary object detection via vision and language knowledge distillation. In: ICLR (2021)

\bibitem{gupta20lvis}
Gupta, A., Dollar, P., Girshick, R.: {LVIS}: A dataset for large vocabulary instance segmentation. In: CVPR (2019)

\bibitem{hu2024dac}
Hu, Z., Sun, Y., Wang, J., Yang, Y.: Dac-detr: Divide the attention layers and conquer. Advances in Neural Information Processing Systems  \textbf{36} (2024)

\bibitem{openclip}
Ilharco, G., Wortsman, M., Wightman, R., Gordon, C., Carlini, N., Taori, R., Dave, A., Shankar, V., Namkoong, H., Miller, J., Hajishirzi, H., Farhadi, A., Schmidt, L.: Openclip (Jul 2021). \doi{10.5281/zenodo.5143773}, \url{https://doi.org/10.5281/zenodo.5143773}, if you use this software, please cite it as below.

\bibitem{kim2023contrastivecfm}
Kim, D., Angelova, A., Kuo, W.: Contrastive feature masking open-vocabulary vision transformer (2023)

\bibitem{rovit2023}
Kim, D., Angelova, A., Kuo, W.: Region-aware pretraining for open-vocabulary object detection with vision transformers. In: CVPR. pp. 11144--11154 (2023)

\bibitem{fvlm}
Kuo, W., Cui, Y., Gu, X., Piergiovanni, A., Angelova, A.: Open-vocabulary object detection upon frozen vision and language models. In: ICLR (2023), \url{https://openreview.net/forum?id=MIMwy4kh9lf}

\bibitem{li2021glip}
Li*, L.H., Zhang*, P., Zhang*, H., Yang, J., Li, C., Zhong, Y., Wang, L., Yuan, L., Zhang, L., Hwang, J.N., Chang, K.W., Gao, J.: Grounded language-image pre-training. In: CVPR (2022)

\bibitem{groundingdino}
Liu, S., Zeng, Z., Ren, T., Li, F., Zhang, H., Yang, J., Li, C., Yang, J., Su, H., Zhu, J., et~al.: Grounding dino: Marrying dino with grounded pre-training for open-set object detection. arXiv preprint arXiv:2303.05499  (2023)

\bibitem{liu2022convnet}
Liu, Z., Mao, H., Wu, C.Y., Feichtenhofer, C., Darrell, T., Xie, S.: A convnet for the 2020s. CVPR  (2022)

\bibitem{ma2023codet}
Ma, C., Jiang, Y., Wen, X., Yuan, Z., Qi, X.: Codet: Co-occurrence guided region-word alignment for open-vocabulary object detection. In: NIPS (2023)

\bibitem{owlvitv22023scaling}
Matthias~Minderer, Alexey~Gritsenko, N.H.: Scaling open-vocabulary object detection. NeurIPS  (2023)

\bibitem{menon2022visual}
Menon, S., Vondrick, C.: Visual classification via description from large language models. ICLR  (2023)

\bibitem{owlvit2023}
Minderer, M., Gritsenko, A., Stone, A., Neumann, M., Weissenborn, D., Dosovitskiy, A., Mahendran, A., Arnab, A., Dehghani, M., Shen, Z., et~al.: Simple open-vocabulary object detection. In: ECCV. pp. 728--755. Springer (2022)

\bibitem{pennington2014glove}
Pennington, J., Socher, R., Manning, C.D.: {GloVe}: Global vectors for word representation. In: EMNLP (2014)

\bibitem{pratt2023doescupl}
Pratt, S., Covert, I., Liu, R., Farhadi, A.: What does a platypus look like? generating customized prompts for zero-shot image classification. In: Proceedings of the IEEE/CVF International Conference on Computer Vision. pp. 15691--15701 (2023)

\bibitem{clip}
Radford, A., Kim, J.W., Hallacy, C., Ramesh, A., Goh, G., Agarwal, S., Sastry, G., Askell, A., Mishkin, P., Clark, J., et~al.: Learning transferable visual models from natural language supervision. In: ICML. pp. 8748--8763. PMLR (2021)

\bibitem{Hanoona2022Bridging}
Rasheed, H., Maaz, M., Khattak, M.U., Khan, S., Khan, F.S.: Bridging the gap between object and image-level representations for open-vocabulary detection. In: NIPS (2022)

\bibitem{ren2023detrex}
Ren, T., Liu, S., Li, F., Zhang, H., Zeng, A., Yang, J., Liao, X., Jia, D., Li, H., Cao, H., Wang, J., Zeng, Z., Qi, X., Yuan, Y., Yang, J., Zhang, L.: detrex: Benchmarking detection transformers (2023)

\bibitem{CC}
Sharma, P., Ding, N., Goodman, S., Soricut, R.: Conceptual captions: A cleaned, hypernymed, image alt-text dataset for automatic image captioning. In: Proceedings of the 56th Annual Meeting of the Association for Computational Linguistics (Volume 1: Long Papers). pp. 2556--2565 (2018)

\bibitem{Shi_2023_ICCV}
Shi, C., Yang, S.: Edadet: Open-vocabulary object detection using early dense alignment. In: Proceedings of the IEEE/CVF International Conference on Computer Vision (ICCV) (October 2023)

\bibitem{zhao2020gtnet}
Shizhen, Z., Changxin, G., Yuanjie, S., Lerenhan, L., Changqian, Y., Zhong, J., Nong, S.: Gtnet: Generative transfer network for zero-shot object detection. In: AAAI (2020)

\bibitem{INSTRUCTOR}
Su, H., Shi, W., Kasai, J., Wang, Y., Hu, Y., Ostendorf, M., Yih, W.t., Smith, N.A., Zettlemoyer, L., Yu, T.: One embedder, any task: Instruction-finetuned text embeddings (2022), \url{https://arxiv.org/abs/2212.09741}

\bibitem{wang2023opencorpus}
Wang, J., Zhang, H., Hong, H., Jin, X., He, Y., Xue, H., Zhao, Z.: Open-vocabulary object detection with an open corpus. In: ICCV. pp. 6759--6769 (2023)

\bibitem{OADP}
Wang, L., Liu, Y., Du, P., Ding, Z., Liao, Y., Qi, Q., Chen, B., Liu, S.: Object-aware distillation pyramid for open-vocabulary object detection. CVPR  (2023)

\bibitem{condhead}
Wang, T.: Learning to detect and segment for open vocabulary object detection. In: CVPR. pp. 7051--7060 (2023)

\bibitem{wu2023baron}
Wu, S., Zhang, W., Jin, S., Liu, W., Loy, C.C.: Aligning bag of regions for open-vocabulary object detection. In: CVPR (2023)

\bibitem{wu2023cora}
Wu, X., Zhu, F., Zhao, R., Li, H.: Cora: Adapting clip for open-vocabulary detection with region prompting and anchor pre-matching. ArXiv  \textbf{abs/2303.13076} (2023)

\bibitem{yang2023language}
Yang, Y., Panagopoulou, A., Zhou, S., Jin, D., Callison-Burch, C., Yatskar, M.: Language in a bottle: Language model guided concept bottlenecks for interpretable image classification. In: Proceedings of the IEEE/CVF Conference on Computer Vision and Pattern Recognition. pp. 19187--19197 (2023)

\bibitem{yao2023detclipv2}
Yao, L., Han, J., Liang, X., Xu, D., Zhang, W., Li, Z., Xu, H.: Detclipv2: Scalable open-vocabulary object detection pre-training via word-region alignment. In: Proceedings of the IEEE/CVF Conference on Computer Vision and Pattern Recognition. pp. 23497--23506 (2023)

\bibitem{yao2022detclip}
Yao, L., Han, J., Wen, Y., Liang, X., Xu, D., Zhang, W., Li, Z., Xu, C., Xu, H.: Detclip: Dictionary-enriched visual-concept paralleled pre-training for open-world detection. NIPS  \textbf{35},  9125--9138 (2022)

\bibitem{zang2022open}
Zang, Y., Li, W., Zhou, K., Huang, C., Loy, C.C.: Open-vocabulary detr with conditional matching (2022)

\bibitem{ovr-cnn}
Zareian, A., Rosa, K.D., Hu, D.H., Chang, S.F.: Open-vocabulary object detection using captions. In: CVPR. pp. 14393--14402 (2021)

\bibitem{zhang2022dino}
Zhang, H., Li, F., Liu, S., Zhang, L., Su, H., Zhu, J., Ni, L.M., Shum, H.Y.: Dino: Detr with improved denoising anchor boxes for end-to-end object detection (2022)

\bibitem{zhao2024ms}
Zhao, C., Sun, Y., Wang, W., Chen, Q., Ding, E., Yang, Y., Wang, J.: Ms-detr: Efficient detr training with mixed supervision. In: Proceedings of the IEEE/CVF Conference on Computer Vision and Pattern Recognition. pp. 17027--17036 (2024)

\bibitem{vlplm2022}
Zhao, S., Zhang, Z., Schulter, S., Zhao, L., Vijay~Kumar, B., Stathopoulos, A., Chandraker, M., Metaxas, D.N.: Exploiting unlabeled data with vision and language models for object detection. In: ECCV. pp. 159--175. Springer (2022)

\bibitem{regionclip2022}
Zhong, Y., Yang, J., Zhang, P., Li, C., Codella, N., Li, L.H., Zhou, L., Dai, X., Yuan, L., Li, Y., et~al.: Regionclip: Region-based language-image pretraining. In: CVPR. pp. 16793--16803 (2022)

\bibitem{detic2022}
Zhou, X., Girdhar, R., Joulin, A., Kr{\"a}henb{\"u}hl, P., Misra, I.: Detecting twenty-thousand classes using image-level supervision. In: ECCV. pp. 350--368. Springer (2022)

\bibitem{zhou2021probablistic}
Zhou, X., Koltun, V., Kr{\"a}henb{\"u}hl, P.: Probabilistic two-stage detection. In: arXiv preprint arXiv:2103.07461 (2021)

\bibitem{zhu2020don}
Zhu, P., Wang, H., Saligrama, V.: {Don't Even Look Once}: Synthesizing features for zero-shot detection. In: CVPR (2020)

\end{thebibliography}
